\documentclass{article}
\usepackage{spconfa4,epsfig}
\usepackage{amssymb}
\usepackage{booktabs}

\usepackage[ruled,vlined,linesnumbered]{algorithm2e}
\usepackage{algpseudocode}
\usepackage{amsmath}
\usepackage{color,colortbl}
\usepackage{hyperref}
\usepackage{enumitem}

\usepackage{booktabs}
\usepackage[capitalize]{cleveref}
\crefname{section}{Sec.}{Secs.}
\Crefname{section}{Section}{Sections}
\Crefname{table}{Table}{Tables}

\let\OLDthebibliography\thebibliography
\renewcommand\thebibliography[1]{
  \OLDthebibliography{#1}
  \setlength{\parskip}{0pt}
  \setlength{\itemsep}{0pt plus 0.3ex}
}

\def\ie{\emph{i.e.}}

\def\wrt{w.r.t. } 
 
\def\etal{\emph{et al.}}
\makeatother

\begin{document}\sloppy

\def\x{{\mathbf x}}
\def\L{{\cal L}}

\title{Feature-Balanced Loss for Long-Tailed Visual Recognition}
%
\name{Mengke Li\quad Yiu-ming Cheung$^{\ast}$ \quad Juyong Jiang
\thanks{$^{\ast}$ Corresponding author. We thank Ms. Hui Ji for her help on part of the experiments and literature review. This work was supported in part by NSFC/RGC JRS Grant: N\_HKBU214/21, GRF Grant: 12201321, NSFC Grant: 61672444, and HKBU Grant: RC-FNRA-IG/18-19/SCI/03. Source code is available at \href{https://github.com/juyongjiang/FBL}{\textcolor{blue}{https://github.com/juyongjiang/FBL}}.}
}

\address{Department of Computer Science, Hong Kong Baptist University, Hong Kong \\
{\tt\small \{csmkli, ymc, csjyjiang\}@comp.hkbu.edu.hk}}
\maketitle

\begin{abstract}
Deep neural networks frequently suffer from performance degradation when the training data is long-tailed because several majority classes dominate the training, resulting in a biased model. Recent studies have made a great effort in solving this issue by obtaining good representations from data space, but few of them pay attention to the influence of feature norm on the predicted results. In this paper, we therefore address the long-tailed problem from feature space and thereby propose the feature-balanced loss. Specifically, we encourage larger feature norms of tail classes by giving them relatively stronger stimuli. Moreover, the stimuli intensity is gradually increased in the way of curriculum learning, which improves the generalization of the tail classes, meanwhile maintaining the performance of the head classes. Extensive experiments on multiple popular long-tailed recognition benchmarks demonstrate that the feature-balanced loss achieves superior performance gains compared with the state-of-the-art methods.
\end{abstract}

\section{Introduction}
\label{sec:intro}

In classification problems, real-world data often exhibits a long-tailed distribution: a few majority classes have large amounts of samples, while numerous minority classes are with only a few samples. This extreme imbalance class distribution leads to the model training dominated by head classes. As a result, the model performance for tail classes is severely degraded. Nowadays, it is still challenging to effectively train a model on long-tailed data in visual recognition tasks.

\begin{figure}[t]
	\centering
	\includegraphics[width=1.0\linewidth]{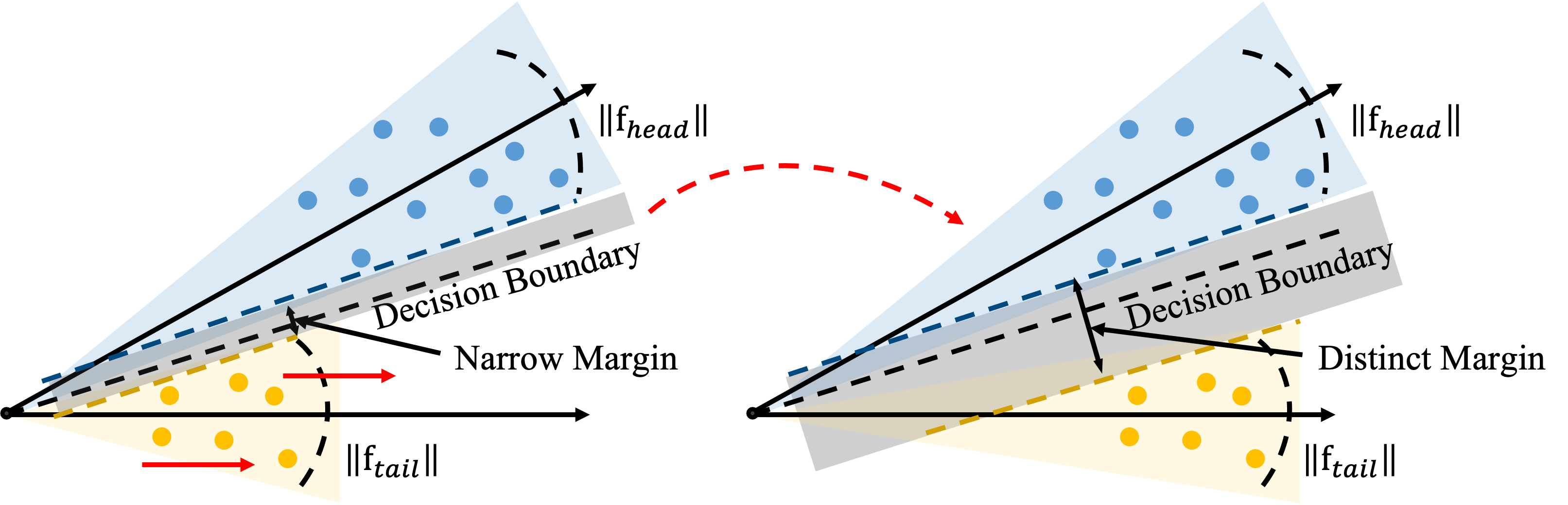}
	\vspace{-12pt}
	\caption{A schematic diagram of the influence of feature norm on decision margin in embedding space. With the increase of the feature norm of the tail class samples, the margin becomes clear and the separability of the samples can be enhanced, which in turn improves the model generalization towards the samples.
	}\label{fig:intro}
	\vspace{-12pt}
\end{figure}

To address the issue of extreme data imbalance caused by long-tailed distribution, an intuitive way is to re-balance the model via class-balanced sampling~\cite{he2008adasyn, kang2020decoupling} or loss function re-weighting~\cite{Huang2016CVPR, Salman2018Cost}. However, these methods result in overfitting to the tail classes, which invariably inhibits the performance of the model. Most recently, Cui~\etal~\cite{cui2019class} have proposed to re-weight the loss function or re-sample the data based on the ``effective number'' of each class, which has been shown empirically effective. This ``effective number'' strategy, on the other hand, does not truly address the issue of uneven feature distribution for long-tailed data. Subsequently, Cao~\etal~\cite{cao2019learning} utilized the label-distribution-aware margin (LDAM) to re-weight the loss, which can improve the generalization performance of tail classes. Nevertheless, it calculates the predicted logit through the cosine distance, which neglects the significant influence of the feature norm.

In this paper, we address the long-tailed problem from a feature norm perspective and thereby proposing the feature-balanced loss (FBL). As shown in~\cref{fig:intro}, it can be seen that training samples with less feature norm are difficult to classify because of the unclear margins between each class. The increase of feature norm can enlarge the margins between classes and enhance the separability of the samples. Based on this observation, we add a class-based stimulus to the predicted logit to encourage larger tail class feature norm to improve its generalization. Different from LDAM that utilizes hard margins to increase intra-class compactness, our FBL enlarges decision margin without compressing the embedding space distribution of each class. Furthermore, we adopt curriculum learning~\cite{bengio2009curriculum} strategy to gradually increase the class-based stimulus so that the network initially concentrates on the head classes, and then gradually shifts its attention to the tail classes as the training progresses. In this way, the classification accuracy of the tail classes can be improved while maintaining the performance of the head classes. We validate the proposed FBL on five popular benchmark datasets, \ie, CIFAT-10-LT, CIFAT-100-LT, ImageNet-LT, iNaturalist 2018 and Places-LT. We also conduct an additional experiment on feature norm visualization, which demonstrates that feature norm is one of the key factors for improving the accuracy of long-tailed data classification.

Our main contributions are summarized as follows:
\begin{itemize}[leftmargin=*]
    \item We propose the novel FBL for long-tailed visual recognition by adding an extra classes-based stimulus to the logit. The proposed FBL encourages larger feature norms for tail classes, thereby improving the generalization performance of these classes.
    \item We propose to gradually increase the intensity of stimulus in the way of curriculum learning. This robust training strategy not only enhances the classification accuracy of tail classes to a large extent, but also maintains the performance of head classes.
    \item We conduct extensive experiments on commonly used long-tailed datasets, which demonstrates the superiority of the proposed method in comparison with the state-of-the-art methods.
\end{itemize}

\section{Related Work}
\label{sec:work}
Long-tailed visual recognition has received increasing attention in computer vision because of the prevalence of data imbalance in the real world. This section will make an overview of the most related works.

\subsection{Loss Modification}
Loss modification aims to re-balance the importance of different classes by tuning the loss values. It addresses the class imbalance problem from two perspectives: sample-wise and class-wise. Sample-wise methods~\cite{Mengye2018Learning, Tsung2020Focal} assign large relative weights to the difficult samples through the fine-grained parameters in the loss. For example, focal loss~\cite{Tsung2020Focal} utilizes the sample prediction hardness as the re-weighting coefficient of the loss function. However, the classification difficulty of a sample may not be directly related to its corresponding class size. Hence, the sample-wise method is incapable of handling the large-scale and severe imbalance data. Class-wise methods~\cite{Salman2018Cost, cui2019class, tan2020equalization} assign the loss function with class-specific parameters that are negatively correlated to the label frequencies. For example, Cui \etal~\cite{cui2019class} proposed to re-weight the loss function by the ``effective number'' of each class instead of label frequency. Nevertheless, it does not completely alleviate the problem of biased feature distribution.

\subsection{Logit Adjustment}
Logit adjustment addresses the class imbalance problem by calibrating the logit to the prior during inference or training. Typically, a number of approaches adjust the loss during training. Most recently, Cao~\etal~\cite{cao2019learning} have proposed label-distribution-aware margin accompanied with the deferred scheme (LDAM-DRW), which enforces tail classes to have large relative margins to increase their classification accuracy. Furthermore, DisAlign~\cite{zhang2021distribution} adaptively aligns the logit to a balanced class distribution to adjust the biased decision boundary, which can re-balance the classifier well. Besides, another kind of method post-hoc shifts the predicted logits. For example, Menon \etal~\cite{menon21long} proposed logit adjustment (LA) to post-process the logit based on the label frequencies of training data. In contrast, Hong~\etal~\cite{Hong2021CVPR} proposed LADE, which post-adjusts logits with the label frequencies of testing data, allowing the distribution of the test set to be arbitrary.

\section{Proposed Method}
\label{sec:method}
To mitigate the training bias towards the head classes caused by long-tailed data, we propose the FBL as a more powerful supervised signal for optimizing deep neural networks (DNNs).

\subsection{Analysis of Softmax Loss Function in Classification}\label{sec:analysis}
Given a training sample $x$ with the label $y$ from the training set $\mathcal{T}$ with total $C$ classes and $N$ training samples. We use $\mathbf{f} \in \mathbb{R}^D$ to represent the feature of $x$ obtained from the embedding layer with dimension $D$. $\mathbf{W}=\{\mathbf{w}_1, \mathbf{w}_2,\cdots,\mathbf{w}_C\} \in \mathbb{R}^{D\times C}$ represents the weight matrix of the classifier, where $\mathbf{w}_i$ represents the weight vector of class $i$ in the classifier. The predicted logit of class $i$ is represented by $z_i$, thus, $z_i = \mathbf{w}_{i}^T \mathbf{f}$. We use the subscript $y$ to represent the target class. That is, $z_{y}$ indicates the target logit and $z_{i}\left(i\neq y\right)$ is the non-target logit. The original softmax loss function for the given sample $x$ is:
\begin{equation}\label{eq:softmax}
  L_\text{softmax}(x) = -\log \frac{e^{z_{y}}}{ \sum_j e^{z_j}}.
\end{equation}
The gradient of $L_\text{softmax}$ \wrt $z_i$ is:
\begin{equation}\label{eq:partial_sm}
\frac{\partial L_\text{softmax}}{\partial z_i} =
\left\{
\begin{array}{lr}
p_i-1, &i = y\\
p_i,   &i\neq y
\end{array},
\right.
\end{equation}
where $p_i = \frac{e^{z_{i}}}{ \sum_j e^{z_j}}$. In backward propagation, the gradients of the target class are negative, and those of the non-target classes are positive. Thus, the training samples punish the non-target class weights $\mathbf{w}_i \left(i \neq y\right)$ by $p_i$. The weights of tail classes which have fewer training instances always receive punishment signals. As a result, the weight norm of the classifier for tail classes is always reduced. Therefore, we obtain the following properties:


\noindent\textbf{Property 1.} The weight norm $\|\textbf{w}_i\|$ of the classifier for class $i$ is correlated with the class size $n_i$.

In addition, we introduce additional property of softmax loss that was found by Yuan~\etal~\cite{Yuan18feature}:

\noindent \textbf{Property 2.} By fixing the weight vector and direction of feature vectors, softmax loss is a function that monotonically decreases with the increasing of feature $L_2$-norm when features are correctly classified.

Property 1 indicates that the target logit $z_y = \mathbf{w}_y^T\textbf{f}$ of tail class is usually suppressed because of the relatively small $\mathbf{w}_y^T$.
Meanwhile, Property 2 shows that feature norm is an important factor to achieve a lower loss, so that the features can be more separable. To improve the performance on tail classes, we can encourage larger feature norm for tail classes to diminish the bias towards the head classes.

\subsection{FBL with Curriculum Learning}
To stimulate the large feature norm, we can add an additional constraint item to the original cross-entropy loss:
\begin{equation}\label{FCL}
  L' = -\log \frac{e^{z_{y}}}{ \sum_i e^{z_i}} + \alpha \frac{\lambda_y}{\mathbf{\|f\|}},
\end{equation}
where $\alpha$ is the parameter used to adjust the strength of the constraint, and $\lambda_y$ controls the stimulus intensity towards different classes. Since Property 1 in \cref{sec:analysis} states that the weight norm of classifier for tail classes is usually suppressed, the logits of tail classes will be unfairly reduced. To diminish this bias, we can encourage large feature norms for tail classes and thus assign them stronger stimulation. Therefore, $\lambda_y$ is negatively correlated with the number of samples in class $y$.

For the sake of analysis of the loss function, we rewrite \cref{FCL} as:
\begin{equation}\label{FCL2}
\begin{array}{lll}
  L' & = -\log \frac{e^{z_{y}}}{ \sum_j e^{z_j}} + \log e^{\frac{\lambda_y}{\mathbf{\|f\|}}} \\
             & = -\log \frac{e^{z_{y}- \frac{\lambda_y}{\mathbf{\|f\|}} }}{ \sum_j e^{z_j}} \\
             & = -\log p_y
\end{array},
\end{equation}
where $p_y = \frac{e^{z_{y}- \frac{\lambda_y}{\mathbf{\|f\|}} }}{ \sum_j e^{z_j}}$. As the sum of the probabilities of all classes obtained by \cref{FCL2} is not equal to 1, \ie, $\sum_{y=1}^C p_y \neq 1$, we further modify the logit to ensure that the total predicted probabilities of all classes are equal to 1. The feature-balanced logit $z^b_j$ of class $j$ is introduced and is expressed as:
\begin{equation}\label{logit}
 z^{b}_j = z_j-\alpha \frac{\lambda_j}{\|\textbf{f}\|}.
\end{equation}
In addition, $\lambda_j$ controls the intensity of the stimulus, which should be weak for head classes and strong for tail classes. Subsequently, we set $\lambda_j$ at:
\begin{equation}\label{lambde}
 \lambda_j = \log n_{max} - \log n_{j},
\end{equation}
so that it is zero for the most frequent class and is much stronger for tail classes.

Furthermore, the stronger the constraint on feature (\ie, $ \frac{\lambda_j}{\|\textbf{f}\|}$) is, the more the model focuses on the tail classes. We can adopt the idea of curriculum learning~\cite{bengio2009curriculum}, which makes the model initially focus on easy samples (\ie, head classes), and then gradually shift to learning difficult samples (\ie, tail classes). To achieve this, we can choose the learning strategy that gradually increases $\alpha$ as the training progresses. Therefore, we replace $\alpha$ by $\alpha(t)$ which is related to the training epoch $t$. We empirically select the parabolic increase learning strategy, which is expressed as:
\begin{equation}\label{alpha}
 \alpha(t) \propto (\frac{t}{T})^2,
\end{equation}
where $t$ is the training epoch and $T$ is the total number of epochs. \cref{sec:curr} also provides experimental results for different learning strategies.

The final loss function $L_\text{FBL}$ is expressed as:
\begin{equation}\label{FBL}
  L_\text{FBL} = -\frac{1}{N}\sum_i\log \frac{e^{z^{b}_{y_i}}}{\sum_j e^{z^{b}_j}}.
\end{equation}
This loss function is named as FBL--feature-balanced loss, because it balances the logit of different classes based on feature norm. The algorithm of our proposed method is summarized in \cref{algorithm}.

\begin{algorithm}[t]
\caption{FBL with curriculum learning}\label{algorithm}
\SetAlgoLined
\KwIn {Training dataset $\mathcal{S}$}
\KwOut {Predicted labels}
Initialize the DNN model $\phi((x,y);\theta)$ randomly, where $\theta$ is the parameter of the model\;
\For {$t=1$ to $T$}{
Sample mini-batch training samples $\mathcal{B}$ from the long-tailed data $\mathcal{S}$ with batch size of $b$\;
Obtain the constraint strength parameter $\alpha$: $\alpha \leftarrow \alpha(t)$\;
Obtain the stimulus intensity parameter $\lambda_j$: $\lambda_j \leftarrow  \log n_{max}-\log n_j$\;
Calculate the loss by \cref{FBL}: $\mathcal{L}((x,y);\theta) = \frac{1}{b}\sum_{(x,y)\in \mathcal{B}} L_\text{FBL}(x,y)$\;
Update model parameters: $\theta \leftarrow \theta - \alpha' \nabla_{\theta} \mathcal{L}((x,y);\theta) $\;}
\end{algorithm}

\section{Experiments}
\label{sec:experiment}
\begin{table*}
\centering  %
	\caption{An Overview of Long-Tailed Datasets}
	\label{Dataset}
    \setlength{\tabcolsep}{8pt}
    \resizebox{0.85\textwidth}{!}
	{
	\begin{tabular}{c|c|c|c|c|c|c|c}
		\toprule[0.8pt]
		Dataset & \multicolumn{2}{c|}{CIFAR-10-LT} & \multicolumn{2}{c|}{CIFAR-100-LT} & ImageNet-LT &Places-LT & iNat 2018 \\
		\hline
        \hline
		 \# Classes  &\multicolumn{2}{c|}{10}&\multicolumn{2}{c|}{100} &1,000 &365 &8,142 \\
		 \hline
		$IF$ &100 & 50 &100 & 50 &256 &996 &500 \\
		\hline
		\# Train. img. &12,406 & 13,996 &10,847 & 12,608 &115,846 &62,500 &437,513  \\
		Tail class size &50 & 100 &5 & 10 &5 &5 &2  \\
		Head class size &5,000 & 5,000 & 500 & 500 &1,280 &4,980 &1,000 \\
		\# Val. img. & - & - & - & - & 20,000 & 7,300 &24,426    \\
	    \# Test img. &10,000 & 10,000 &10,000 &10,000 &50,000 &36,500 &- \\
		\bottomrule[0.8pt]
	\end{tabular}
	}
\vspace{-6pt}
\end{table*}
\subsection{Datasets}
To demonstrate the effectiveness of our proposed FBL, we conduct the experiments on five benchmark datasets with the various scales.

\noindent\textbf{CIFAR-10/100-LT}~\cite{cao2019learning} down-samples the original balanced version of CIFAR-10/100~\cite{krizhevsky2009learning} per class by an imbalanced factor $IF=\frac{N_{max}}{N_{min}}$ (where $N_{max}$ and $N_{min}$ are the numbers of training samples in the most and the least frequent classes, respectively). CIFAR-10-LT and CIFAR-100-LT have two typical variants, namely, with $IF=\{100, 50\}$.

\noindent\textbf{ImageNet-LT}~\cite{liu2019large} is a large-scale long-tailed dataset for object classification through sampling a subset following the Pareto distribution with the power value $\alpha=6$ from ImageNet-2012~\cite{Deng_2009_CVPR}. It includes 115.8K images with the class size ranging from 5 to 1,280, imitating the long-tailed distribution that regularly existed in the real world.

\noindent\textbf{Places-LT} is a long-tailed version of the large-scale scene classification dataset Places-365~\cite{Zhou2018places}. There are 184.5K images with class sizes ranging from 5 to 4,980. Moreover, the gap between the sizes of tail and head classes of this dataset is larger than that of ImageNet-LT.

\noindent\textbf{iNaturalist 2018 (iNat 2018)} is the iNaturalist species classification and detection dataset~\cite{van2018inaturalist}, which is a massive real-world long-tailed dataset. In its 2018 version, iNaturalist comprises 437,513 training images from 8,142 classes. In the light of different classes, the numbers of the training samples follow an exponential decay.

\cref{Dataset} summarizes the details of the above datasets.

\subsection{Implementation Details}
We use Pytorch to implement and train all the backbones with stochastic gradient descent with momentum. %

\noindent\textbf{Backbone.}
Following the protocol of Cui~\etal~\cite{cui2019class}, ResNet-32 is adopted as the backbone for all CIFAR-10/100-LT datasets. For ImageNet-LT and iNat 2018, ResNet-50 is applied. For Places-LT, we follow Liu \etal~\cite{liu2019large} and start from a ResNet-152 pre-trained on the original balanced version of ImageNet. Except for ResNet-152, all the backbones are trained from scratch.

\noindent\textbf{Training details.}
For CIFAR-10/100-LT, we train the backbone with 200 epochs and batch size of 64. The initial learning rate ($lr$) is set at $0.1$, and we anneal $lr$ by 100 at the 160-th and 180-th epoch, respectively. For the three large-scale datasets, backbone is trained with 180 epochs, batch size of 512, and initial $lr=0.2$. We divide $lr$ by 10 at 120-th and 160-th epochs.

\subsection{Comparison Methods}
The vanilla training with cross-entropy (CE) loss is chosen as the baseline method. We compare the proposed method with the state-of-the-art ones, \ie, the logit modification methods including: LDAM-DRW~\cite{cao2019learning} and LA~\cite{menon21long}, the most recently proposed two-stage method--BBN~\cite{bbn20} on the small-scale datasets (CIFAR-10/100-LT) and decoupling on the large-scale datasets (imageNet-LT, iNat 2018 and Places-LT).

\subsection{Long-Tailed Recognition Results}
\begin{table}[t]
\centering
    \caption{Comparison results on CIFAR-10/100-LT. Top-1 accuracy (\%) are reported. The best results are shown in \underline{\textbf{underline bold}}.}
    \label{cifar_results}
    \resizebox{0.47\textwidth}{!}{
      \begin{tabular}{c |c c | c c}
      \toprule[0.8pt]
      Dataset & \multicolumn{2}{c|}{CIFAR-10-LT} & \multicolumn{2}{c}{CIFAR-100-LT}\\ \hline
      \hline
      Backbone Net& \multicolumn{4}{c}{ResNet-32}\\ %
      \hline		
      \emph{IF} &100 &50  &100 &50 \\
      \hline
      CE loss (baseline) &71.07  &75.31  &39.43 &44.20\\
      \hline
      LDAM-DRW \cite{cao2019learning} (\textit{NeurIPS} 2019)  &77.03  &81.03 &42.04 &47.62\\
      BBN~\cite{bbn20} (\textit{CVPR} 2020) &79.82 &81.18 &42.56 &47.02\\
      LA \cite{menon21long}(\textit{ICLR} 2021)&80.92&- &43.89 &-\\
      \hline
      {\textbf{FBL (ours)}} &\underline{\textbf{82.46}} &\underline{\textbf{84.30}}  &\underline{\textbf{45.22}} &\underline{\textbf{50.65}} \\
      \bottomrule[0.8pt]
     \end{tabular}
    }
    \vspace{-15pt}
\end{table}
\begin{table}[t]
\centering
    \caption{Comparison results on ImageNet-LT, iNaturalist 2018 and Places-LT. Top-1 accuracy (\%) are reported. The best results are shown in \underline{\textbf{underline bold}}.}
    \label{large_dataset_results}
    \resizebox{0.47\textwidth}{!}{
      \begin{tabular}{c |c|c|c }
      \toprule[0.8pt]
      Dataset & ImageNet-LT & iNat 2018 & Places-LT\\ \hline
      \hline
      Backbone Net & ResNet-50 & ResNet-50  & ResNet-152  \\ %
      \hline
      CE loss (baseline) &44.51 &63.80   & 27.13 \\
      \hline
      LDAM-DRW \cite{cao2019learning} (\textit{NeurIPS} 2019) &48.80  &68.00 &-  \\
      Decoupling~\cite{kang2020decoupling} (\textit{ICLR} 2020) &47.70  &69.49 &37.62  \\
      LA \cite{menon21long} (\textit{ICLR} 2021) &50.44 & 66.36 &- \\
      \hline
      {\textbf{FBL (ours)}} &\underline{\textbf{50.70}} &\underline{\textbf{69.90}} &\underline{\textbf{38.66}} \\
      \bottomrule[0.8pt]
     \end{tabular}
    }
    \vspace{-6pt}
\end{table}

\begin{table}[t]
 \centering  
 \caption{Ablation experiment of different learning strategy on CIFAR-10-LT with $IF = 100$.} \label{ab_rs}
 \resizebox{0.38\textwidth}{!}
 {\begin{tabular}{c |c |c}  
  \toprule[0.6pt]
  $\alpha(t) $ & Representation & Acc.(\%)  \\
  \hline
  \hline
        Linear decrease & $1-t/T$ & 75.97 \\
        Linear increase & $t/T$ &  81.67 \\
        Sine increase   & $\sin (t/T \cdot \pi/2)$ & 81.22 \\
        Cosine increase & $1-\cos (t/T \cdot \pi/2)$ & 80.79 \\
        Parabolic increase & $(t/T)^2$ & \underline{\textbf{82.46}} \\
  \bottomrule[0.6pt]
 \end{tabular}}
\vspace{-12pt}
\end{table}

\begin{figure*}[t]
	\centering
	\includegraphics[width=0.98\linewidth]{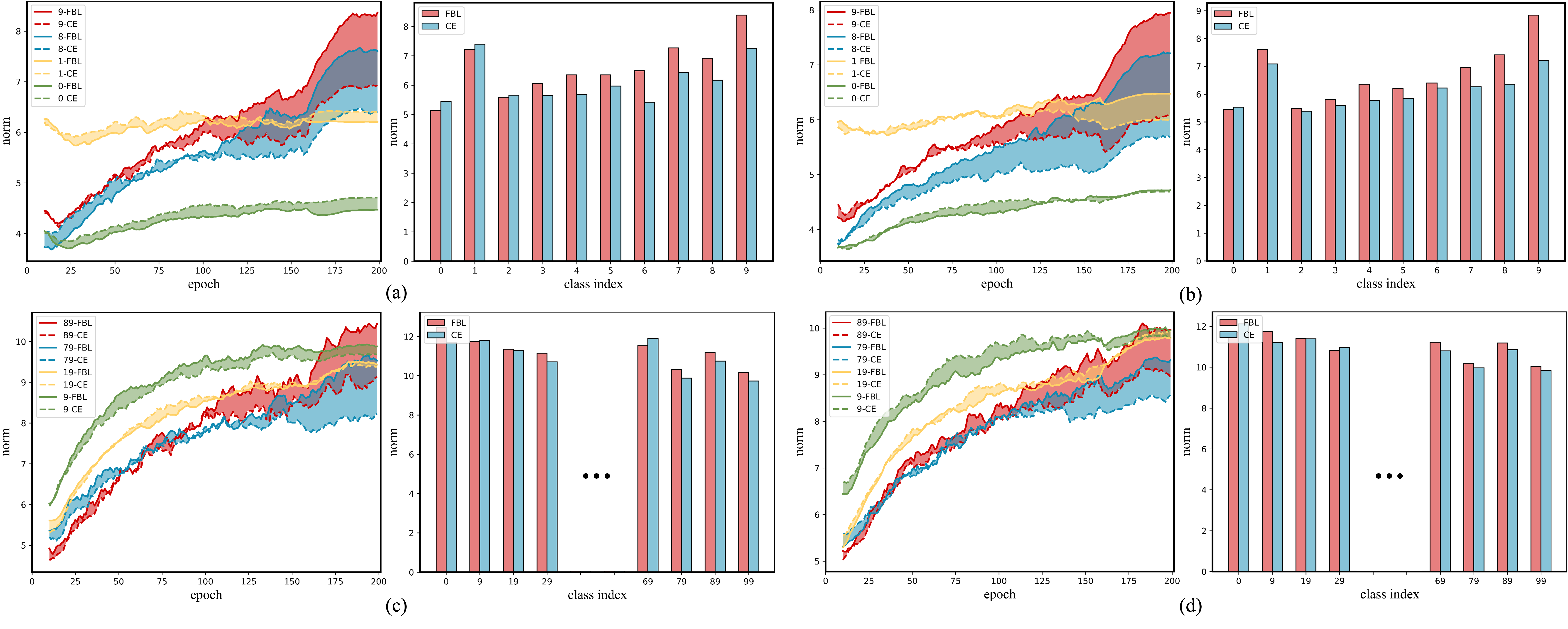}
	\vspace{-12pt}
	\caption{\textbf{Top}: The changes of feature norm on \emph{head classes} (class index-$\{0, 1\}$) and \emph{tail classes} (class index-$\{8, 9\}$) with respect to training epochs (left) and the feature norm distribution of classes over test dataset (right) on CIFAR-10 with $IF = 100$ (a) and 50 (b). \textbf{Bottom}: The changes of feature norm on \emph{head classes} (class index-$\{9, 19\}$) and \emph{tail classes} (class index-$\{79, 89\}$) with respect to training epochs (left) and the feature norm distribution of classes over test dataset (right) on CIFAR-100 with $IF = 100$ (c) and 50 (d).
	}
	\label{fig:vis}
	\vspace{-12pt}
\end{figure*}

\textbf{Results on CIFAR-10/100-LT.} We conduct the comparison experiments on CIFAR-10/100-LT with $IF=\{100, 50$\}. \cref{cifar_results} summarizes the top-1 accuracy. Our FBL outperforms the other competing methods by noticeable margins across all the datasets. For example, FBL outperforms the state-of-the-art method -- LA by 1.54$\%$ and 1.33$\%$ with $IF=100$ on CIFAR-10-LT and CIFAR-100-LT, respectively.

\noindent\textbf{Results on large-scale datasets.}
FBL yields good performance on all large-scale datasets, which is consistent with that CIFAR-10/100-LT.  \cref{large_dataset_results} shows the comparison results. The proposed FBL that can be trained end-to-end not only achieves better results than LA, but also is superior to the two-stage method, \ie, LDAM-DRW and decoupling. For example, on ImageNet-LT, FBL outperforms LDAM-DRW and decoupling by $1.90\%$ and $3.00\%$, respectively.

\subsection{Ablation Study}\label{sec:curr}

We conduct an ablation study to investigate the effectiveness of different learning strategies adopted by $\alpha(t)$. \cref{ab_rs} summarizes their performance. It can be seen that the classification accuracy of the linear decrease strategy is 75.97$\%$, which is only higher than that of the baseline method (71.07$\%$). It is not as competitive as other learning strategies, because it makes the DNN model focus on hard samples (\ie, tail classes) first. As the training progresses, the network gradually forgets what it has previously learned. Therefore, there is basically no improvement in the performance of the tail classes. Other strategies that increase $\alpha$ with the training epoch $t$ gradually shift the network's attention from the head classes to the tail, which can avoid forgetting the tail classes and improve the overall performance.

\begin{table}[t]
 \begin{minipage}[t]{0.5\textwidth}
 \centering  
	\caption{Per-class accuracy ($\%$) of test set on CIFAR-10-LT.}
	\label{tab:per-class-acc-10}
    \setlength{\tabcolsep}{8pt}
    \resizebox{0.95\textwidth}{!}
	{
	\begin{tabular}{c |c c c c c c c c c c}
		\toprule[0.8pt]
		Class index & 0 & 1 & 2& 3& 4& 5& 6& 7& 8& 9 \\
		\hline
        \hline
		$IF$ & \multicolumn{10}{c}{100} \\
        \hline		
		CE loss & \underline{\textbf{91.0}}& \underline{\textbf{98.2}}& \underline{\textbf{83.2}}& 72.5& 78.8& 65.1& 68.8& 59.5& 49.0& 44.6\\
		FBL & 88.1& 94.7& 81.9& \underline{\textbf{73.0}}& \underline{\textbf{83.6}}& \underline{\textbf{75.1}}& \underline{\textbf{86.3}}& \underline{\textbf{77.3}}& \underline{\textbf{82.7}}& \underline{\textbf{81.9}}\\
		\hline
		$IF$ & \multicolumn{10}{c}{50} \\
        \hline		
		CE loss & \underline{\textbf{84.5}}& \underline{\textbf{95.8}}& 68.5& \underline{\textbf{74.6}}& 81.1& 72.7& 82.9& 67,5& 59.1& 66.4 \\
		FBL & \textcolor{black}{83.7}& \textcolor{black}{92.1}& \underline{\textbf{81.7}}& \textcolor{black}{73.9}& \underline{\textbf{85.0}}& \underline{\textbf{76.1}}& \underline{\textbf{87.7}}& \underline{\textbf{85.0}}& \underline{\textbf{88.5}}& \underline{\textbf{89.3}}\\
		\bottomrule[0.8pt]
	\end{tabular}
	}
 \end{minipage}

\begin{minipage}[t]{0.5\textwidth}
\centering  %
	\caption{Per-class accuracy ($\%$) of test set on CIFAR-100-LT.}
	\label{tab:per-class-acc-100}
    \setlength{\tabcolsep}{8pt}
    \resizebox{0.95\textwidth}{!}
	{
	\begin{tabular}{c |c c c c c c c c c c}
		\toprule[0.8pt]
		Class index & 0 & 9 & 19& 29& \multicolumn{2}{c}{$\cdots$} & 69& 79& 89& 99 \\
		\hline
        \hline
		$IF$ & \multicolumn{10}{c}{100} \\
        \hline		
		CE loss & \underline{\textbf{89.0}}& 72.0& \underline{\textbf{59.0}}& \underline{\textbf{48.0}}& \multicolumn{2}{c}{$\cdots$}& 45.0& 12.0& 3.0& 2.0\\
		FBL& 86.0& \underline{\textbf{77.0}}& 54.0& 45.0& \multicolumn{2}{c}{$\cdots$}& \underline{\textbf{60.0}}& \underline{\textbf{27.0}}& \underline{\textbf{22.0}}& \underline{\textbf{8.0}}\\
		\hline
		$IF$ & \multicolumn{10}{c}{50} \\
        \hline		
		CE loss & \underline{\textbf{88.0}}& \underline{\textbf{79.0}}& 53.0& 49.0& \multicolumn{2}{c}{$\cdots$}& 53.0& 10.0& 19.0& 13.0 \\
		FBL & 87.0& 77.0& \underline{\textbf{56.0}}& \underline{\textbf{57.0}}& \multicolumn{2}{c}{$\cdots$}& \underline{\textbf{62.0}}& \underline{\textbf{48.0}}&\underline{\textbf{38.0}}& \underline{\textbf{17.0}}\\
		\bottomrule[0.8pt]
	\end{tabular}
	}
 \end{minipage}
 \vspace{-12pt}
\end{table}

\subsection{Feature-balanced Results}
To further validate the effects of the proposed FBL, especially the tail classes, we visualize the changes of the feature norm (\ie, $\|\textbf{f}\|$) with respect to training epochs and feature norm distribution of classes over the test set on CIFAR-10/100-LT. The results are shown in \cref{fig:vis}. The corresponding per-class accuracy is presented in Table \ref{tab:per-class-acc-10} and \ref{tab:per-class-acc-100}, respectively. The following phenomena can be seen:

\begin{itemize}[leftmargin=*]
    \item The capability of the model to learn from different classes of samples is diverse. Specifically, in \cref{fig:vis} (a) and (b), the feature norms of head classes samples (class index-$\{0, 1\}$) reach stable in very early training epochs due to enough training samples. Differently, on CIFAR-100-LT (as shown in \cref{fig:vis} (c) and (d)), the feature norms of the samples from all classes including head (class index-$\{9, 19\}$) and tail classes (class index-$\{ 79, 89\}$) are constantly changing as the epoch increases, which have a similar phenomenon to the tail classes in CIFAR-10-LT (class index-$\{8, 9\}$ in \cref{fig:vis} (a) and (b)) because they all suffer from insufficient training samples.
    \item Compared with CE loss, our FBL encourages larger feature norms of tail class samples to eliminate representation bias towards head classes. The area $\left(S^{\ast}_{area}(\text{class index})\right)$ enclosed by the curve of CE loss and FBL becomes larger as the number of class samples decreases, e.g. $S^{tail}_{area}(9)>S^{tail}_{area}(8)>S^{head}_{area}(1)>S^{head}_{area}(0)$ in \cref{fig:vis} (a), which is in line with our motivation.
\end{itemize}
These observations not only justify our intuition about the influence of feature norm on decision margin, but also offer a new promising way to investigate long-tailed visual recognition.

\section{Conclusions}
\label{sec:conclusion}
In this work, we have proposed a novel FBL to address the long-tailed classification from feature space. FBL encourages larger feature norms of tail classes by adding relatively stronger stimuli to the logits of tail classes, which can mitigate the representation bias towards head classes in the feature space. In addition, a curriculum learning strategy has been adopted to gradually increase the stimuli in training, which can keep the good accuracy of the model for the head classes and improve the performance of the tail classes. FBL allows DNNs to be trained end-to-end without the risk of a performance drop from head classes. Extensive experiments have demonstrated the superiority of the proposed FBL.

\bibliographystyle{IEEEbib}
\bibliography{ref}

\end{document}